\documentclass[10pt,twocolumn,letterpaper]{article}
\usepackage{cvpr}
\usepackage{times}
\usepackage{epsfig}
\usepackage{graphicx}
\usepackage{amsmath}
\usepackage{amssymb}
\usepackage{pbox}
\usepackage{epstopdf}
\usepackage{subfigure}
\usepackage{xspace}
\usepackage{comment}
\usepackage{lipsum}
\usepackage{booktabs}
\usepackage{bm}
\usepackage{bbm}
\usepackage{tabularx}
\usepackage{setspace}
\usepackage{sidecap}

\usepackage[pagebackref=true,breaklinks=true,letterpaper=true,colorlinks,bookmarks=false]{hyperref}

\cvprfinalcopy 

\def\cvprPaperID{1371} 

\ifcvprfinal\fi

\begin{document}

\title{GSPN: Generative Shape Proposal Network for 3D Instance Segmentation in Point Cloud}

\author{Li Yi$^1$ \qquad Wang Zhao$^2$ \qquad He Wang$^1$ \qquad Minhyuk Sung$^1$  \qquad Leonidas Guibas$^1$\\$^1$Stanford University \qquad $^2$Tsinghua University}


\maketitle
\setlength\abovedisplayskip{5pt}
\setlength\belowdisplayskip{5pt}

\begin{abstract}
We introduce a novel 3D object proposal approach named Generative Shape Proposal Network (GSPN) for instance segmentation in point cloud data. Instead of treating object proposal as a direct bounding box regression problem, we take an analysis-by-synthesis strategy and generate proposals by reconstructing shapes from noisy observations in a scene. We incorporate GSPN into a novel 3D instance segmentation framework named Region-based PointNet (R-PointNet) which allows flexible proposal refinement and instance segmentation generation. We achieve state-of-the-art performance on several 3D instance segmentation tasks. The success of GSPN largely comes from its emphasis on geometric understandings during object proposal, which greatly reducing proposals with low objectness.
\end{abstract}
\vspace{-\baselineskip}

\section{Introduction}
\noindent Instance segmentation is one of the key perception tasks in computer vision, which requires delineating objects of interests in a scene and also classifying the objects into a set of categories. 3D instance segmentation, with a huge amount of applications in robotics and augmented reality, is in tremendous demand these days. However, the progress of 3D instance segmentation lags far behind its 2D counterpart~\cite{he2017mask, liu2018path, li2016fully}, partially because of the expensive computation and memory cost of directly applying 2D Convolutional Neural Networks (CNN) approaches to 3D volumetric data~\cite{song2016deep, deng2017amodal}. Recently, \cite{qi2017pointnet, qi2017pointnet++} proposed efficient and powerful deep architectures to directly process point cloud, which is the most common form of 3D sensor data and is very efficient at capturing details in large scenes. This opens up new opportunities for 3D instance segmentation and motivates us to work with 3D point clouds.

\begin{figure}[t]
    \centering
    \includegraphics[width=\linewidth]{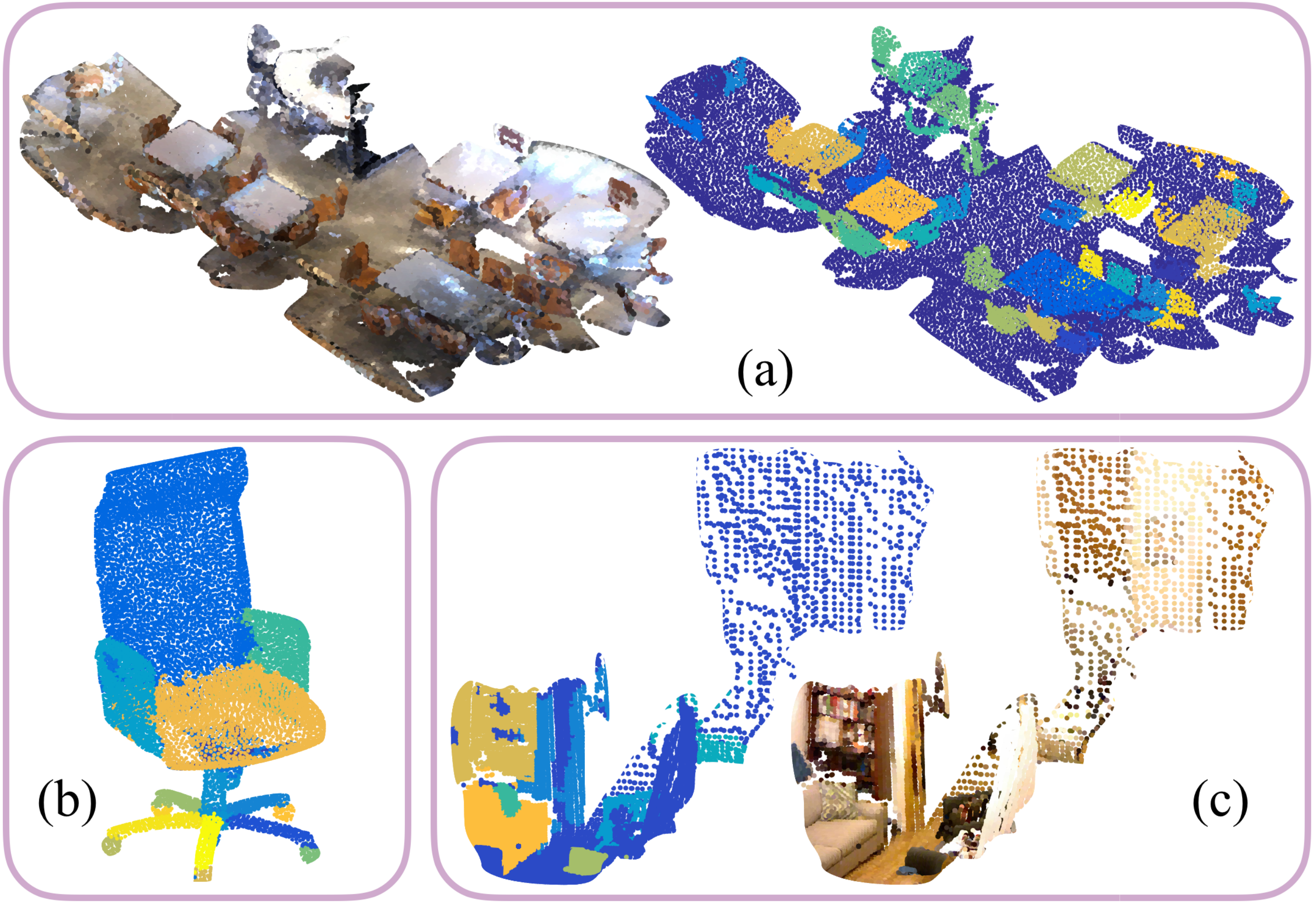}
    \caption{The flexibility of our instance segmentation framework R-PointNet allows it to well handle various types of input data including (a) a complete reconstruction for a real indoor scene, (b) objects with fine-grained part instances, (c) partial point cloud captured from a single view point.}
    \label{fig:teaser}
    \vspace{-1.5\baselineskip}
\end{figure}

The great advance in 2D instance segmentation is largely driven by the success of object proposal methods \cite{uijlings2013selective}, where object proposals are usually represented as 2D bounding boxes.
Thus it is natural to seek for an effective way of representing and generating object proposals in the 3D point cloud. But, this is indeed very challenging since 3D object proposal approaches need to establish the understanding of semantics and objectness for a wide range of object categories with various scales in a cluttered scene under heavy sensor noise and data incompleteness.
A straightforward method is to directly estimate simple geometric approximations to the objects such like 3D bounding boxes \cite{song2016deep,zhou2017voxelnet}. These approximations are simple and not very faithful to most objects, meaning generating such proposals does not require a strong understanding of the underlying object geometry. This makes it easy to produce blind box proposals which either contain multiple objects or just a part of an object, influencing the differentiation among object instances. Moreover, we hardly know how well the 3D object proposer understands objectness.

In contrast, we take a different perspective for object proposal which emphasizes more on the geometric understanding. It is a commonplace that perception is in part constructive \cite{james1890principles}. Therefore we leverage an analysis-by-synthesis strategy where we propose objects by first reconstructing them. Specifically, we leverage a generative model to explicitly depict the natural object distribution and propose candidate objects by drawing samples from the object distribution. The noisy observations in a scene will guide the proposal generation process by indicating where to sample in the object distribution. This idea is especially appealing in 3D since unlike 2D images, objects are more canonical in the 3D world with the right physical scale and more complete geometry. As a result, the object distribution is more compact, making it feasible to capture.

We design a deep neural network named Generated Shape Proposal Network~(GSPN) to achieve this purpose. Compared with direct 3D bounding boxes regression, the advantages of GSPN are two-fold. Firstly, it produces object proposals with higher objectness. The network is explicitly trained to understand how natural objects look like before it generates any proposal. By enforcing geometric understanding, we can greatly reduce blind box proposals not corresponding to a single object. Secondly, GSPN encodes noisy observations to distributions in the natural object space, which can be regarded as an instance-aware feature extraction process. These features delineate object boundaries and could serve as a very important cue for proposal refinement and segmentation mask generation.

To be able to reject, receive and refine object proposals and further segment out various instances in a 3D point cloud, we develop a novel 3D instance segmentation framework called Region-based PointNet (R-PointNet). From a high level, R-PointNet resembles image Mask R-CNN \cite{he2017mask}; it contains object proposal and proposal classification, refinement and segmentation components. We carefully design R-PointNet so that it can nicely consume unstructured point cloud data and make the best use of object proposals and instance sensitive features generated by GSPN.

We have tested our instance segmentation framework R-PointNet with GSPN on various tasks including instance segmentation on complete indoor reconstructions, instance segmentation on partial indoor scenes, and object part instance segmentation. We achieve state-of-the-art performance on all these tasks.

Key contributions of our work are as follows:
\begin{itemize}
\setlength\itemsep{-0.2em}
    \item We propose a Generative Shape Proposal Network to tackle 3D object proposal following an analysis-by-synthesis strategy.
    \item We propose a flexible 3D instance segmentation framework called Region-based PointNet, with which we achieve state-of-the-art performance on several instance segmentation benchmarks.
    \item We conduct extensive evaluation and ablation study to validate our design choices and show the generalizability of our framework.
\end{itemize}

\section{Related Work}
\paragraph{Object Detection and Instance Segmentation}
Recently, a great progress has been made for 2D object detection \cite{girshick2015fast, ren2015faster, redmon2016you, lin2017feature, lin2018focal, liu2016ssd} and instance segmentation \cite{dai2016instance, he2017mask, li2016fully, pinheiro2016learning}. R-CNN~\cite{girshick2014rich} firstly combines region proposal with CNN for 2D object detection. After this, a series of works including Fast R-CNN~\cite{girshick2015fast}, Faster R-CNN~\cite{ren2015faster} and Mask R-CNN~\cite{he2017mask} are proposed to accelerate region proposal, improve feature learning, and extend the detection framework for the instance segmentation task. 

Following these progress in 2D, learning-based 3D detection and instance segmentation frameworks gradually emerged. People largely focused on 3D object bounding box detection where object proposal is essential and have come up with different approaches. \cite{song2016deep} directly apply region proposal network (RPN)~\cite{ren2015faster} on volumetric data, which is limited due to its high memory and computation cost.
\cite{deng2017amodal, qi2017frustum, zhou2017voxelnet, chen2017multi, yang2018pixor, mousavian20173d} rely on the mature 2D object proposal approaches and obtain object proposal from projected views of a 3D scene. It is hard to apply these methods to segment cluttered indoor environment which cannot be fully covered by a few views. In addition, the projection loses information about the scene such as the physical sizes of objects and introduces additional difficulty in object proposal, making it less appealing than directly proposing objects in 3D. 
As a pioneering work in 3D instance segmentation learning, \cite{wang2018sgpn} proposes a shape-centric way to directly propose objects in the 3D point cloud, where points are grouped with a learned similarity metric to form candidate objects. However this bottom-up grouping strategy cannot guarantee proposals with high objectness. In contrast to previous approaches, our approach directly proposes objects in 3D and emphasizes the proposal objectness through a generative model.

\vspace{-\baselineskip}
\paragraph{3D Generative Models}
Variational Autoencoder (VAE) \cite{kingma2013auto} is one of the most popular generative models commonly used for image or shape generation \cite{gulrajani2016pixelvae, nash2017shape}. It learns to encode natural data samples $x$ into a latent distribution where samples can be drawn and decoded to the initial data form. VAE explicitly models the data distribution and learns a proper parametrization via maximizing the data likelihood. However, VAE cannot add controls to the sampled data points, which usually restricts its usage. An extension called Conditional Variational Autoencoder (CVAE) was proposed in \cite{sohn2015learning}, where the generation is also conditioned on certain attributes. 



Alternative to VAE and CVAE, GAN~\cite{goodfellow2014generative} and CGAN~\cite{mirza2014conditional} could generate more faithful images or shapes by introducing an adversarial game between a discriminator and a generator. However, it is well-known that GAN suffers from mode collapse issues since it doesn't explicitly model the data likelihood, while likelihood-based models, such as VAE, can generally capture a more complete data distribution. We leverage CVAE instead CGAN since it complies with the condition more on average.

\vspace{-\baselineskip}
\paragraph{Deep Learning on Point Cloud}
Various 3D representations \cite{su2015multi, wu20153d, riegler2017octnet, yi2017syncspeccnn, qi2017pointnet, qi2017pointnet++, fan2017point} have been explored recently for deep learning on 3D data. Among them, point cloud representation is becoming increasingly popular due to its memory efficiency and intuitiveness. We use some existing deep architectures for 3D point cloud such as PointNet/PointNet++~\cite{qi2017pointnet,qi2017pointnet++} and the Point Set Generation networks~\cite{fan2017point} as the bases of our 3D instance segmentation network.


\section{Method}
\begin{figure*}[t!]
    \centering
    \includegraphics[width=\linewidth]{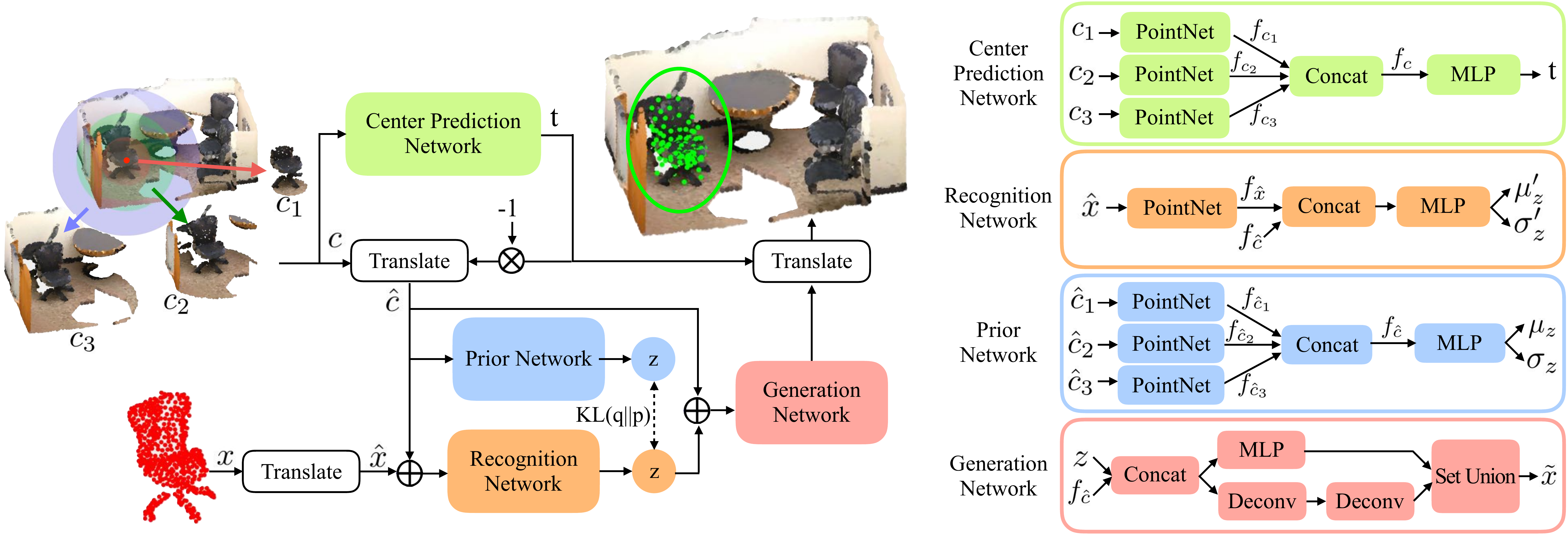}
    \caption{The architecture of GSPN. On the left we show the data flow in GSPN and the key building blocks, highlighted by colored rectangles. The detailed architecture of each building block is shown on the right.}
    \label{fig:gspn}
    \vspace{-\baselineskip}
\end{figure*}

We propose R-PointNet, a 3D object instance segmentation framework that shares a similar high-level structure with 2D Mask R-CNN~\cite{he2017mask} but is carefully designed for unstructured point cloud data. Most importantly, it leverages a network named Generative Shape Proposal Network (GSPN) that efficiently generates 3D object proposals with high objectness. In addition, our Point RoIAlign layer is designed to collect features for proposals, allowing the network to refine the proposals and generate segments. Next, we will explain the design details of our network.


\subsection{Generative Shape Proposal Network}

GSPN treats the object proposal procedure as a object generation, sampling from a conditional probability distribution of natural objects $p_{\theta}(x|c)$ \emph{conditioned} on the scene point cloud $P$ and the seed point $s$, where $c$ represents the context pair $(P, s)$. The output point cloud $\tilde{x}$ generated as an object proposal approximates the object $x$ in $P$ containing $s$.
This approach allows us to concretely see what a certain object proposal looks like and understand whether the network learns the objectness.
Specifically, we formulate GSPN as a conditional variational auto-encoder (CVAE)~\cite{sohn2015learning}.
When approximating $p_{\theta}(x|c)$ as $\int_z p_{\theta}(x|z, c)p_{\theta}(z|c)dz$ with a latent representation $z$ of natural objects, the proposals are generated by drawing a sample $z$ from the conditional prior distribution $p_{\theta}(z|c)$, and then computing object proposals $\tilde{x}$ through the generative distribution $p_{\theta}(x|z, c)$. $p_{\theta}(z|c)$ and $p_{\theta}(x|z, c)$ are learned by maximizing the following
variational lower bound of the training data conditional log-likelihood $\text{log} p_{\theta}(x|c)$:
\begin{equation}
\resizebox{.9\hsize}{!}{$
L = \mathbb{E}_{q_{\phi}(z|x,c)}[\text{log}p_{\theta}(x|z, c)]-\text{KL}(q_{\phi}(z|x, c)||p_{\theta}(z|c))$}
\end{equation}

\noindent
where $q_{\phi}(z|x,c)$ is a proposal distribution which approximates the true posterior $p_{\theta}(z|x,c)$.

The architecture of GSPN is shown in Figure~\ref{fig:gspn}.
Two sub-networks, prior network and recognition network, parameterize $p_{\theta}(z|c)$ and $q_{\phi}(z|x,c)$ as Gaussian distributions with predicted means and variances, respectively, and generation network learns $p_{\theta}(x|z, c)$. In addition, the center prediction network is used to centralize the context data and factor out the location information.
The context pair $c$ is represented by cropping $P$ with spheres centered at $s$ with $K=3$ different radiuses to cover objects with various scales ($c_{k \in \{1 \cdots K\}}$ will denote the context in each scale).
We explain each sub-network next.

Center prediction network takes the context $c$ as input and regresses the center $t$ of the corresponding object $x$ in the world coordinate system (the center is the axis-aligned bounding box center).
The network employs $K$ different PointNets each of which processes a context of each scale $c_k$ and outputs a feature vector $f_{c_k}$ independently, and it concatenates $\{f_{c_k}\}_{k=1}^K$ to form $f_c$, feeds the $f_c$ through a multi-layer perceptron (MLP), and regresses the centroid location $t$.
After this, the context $c$ is centered at $t$, and the translated context $\hat{c}$ serves as the input to the prior network.

The prior network takes the same $K$-PointNet architecture for processing the input centered context $\hat{c}$, and maps the concatenated feature $f_{\hat{c}}$ to a Gaussian prior distribution $\mathcal{N}(\mu_z, \sigma_z^2)$ of $p_{\theta}(z|c)$ through an MLP. Recognition network shares the context encoder with the prior network, and it also consumes an centered object $\hat{x}$ and generates an object feature $f_{\hat{x}}$ with another PointNet. $f_{\hat{x}}$ is then concatenated with the context feature $f_{\hat{c}}$ and fed into an MLP for predicting the Gaussian proposal distribution $\mathcal{N}(\mu_z^{\prime}, \sigma_z^{\prime 2})$, which parametrizes $q_{\phi}(z|x,c)$.

After predicting $p_{\theta}(z|c)$ with the prior network, we sample $z$ and feed it to the generation network. Again, the generation network shares the context encoder with the prior network. After concatenating the context features $f_{\hat{c}}$ from the prior network with $z$, it decodes a point cloud $\tilde{x}$ along with a per-point confidence score $e$ representing the likelihood of appearance for each generated point.
For decoding, we use a point set generation architecture in~\cite{fan2017point} having two parallel branches, fully-connected (fc) branch and deconvolution (deconv) branches, and taking the union of two outputs.
The resulting centralized point cloud is shifted back to its original position with the predicted object center $t$.

GSPN takes an additional MLP taking $f_{\hat{c}}$ predicting objectness score for each proposal similarly with Mask R-CNN.
The objectness score is supervised with axis-aligned bounding boxes; positive proposals come from seed points $s$ belonging to foreground objects, and their bounding boxes overlap with some ground truth boxes with an intersection over union (IoU) larger than $0.5$. Negative proposals are those whose IoUs with all the ground truth boxes are less than $0.5$. 

\begin{figure}
    \centering
    \includegraphics[width=\linewidth]{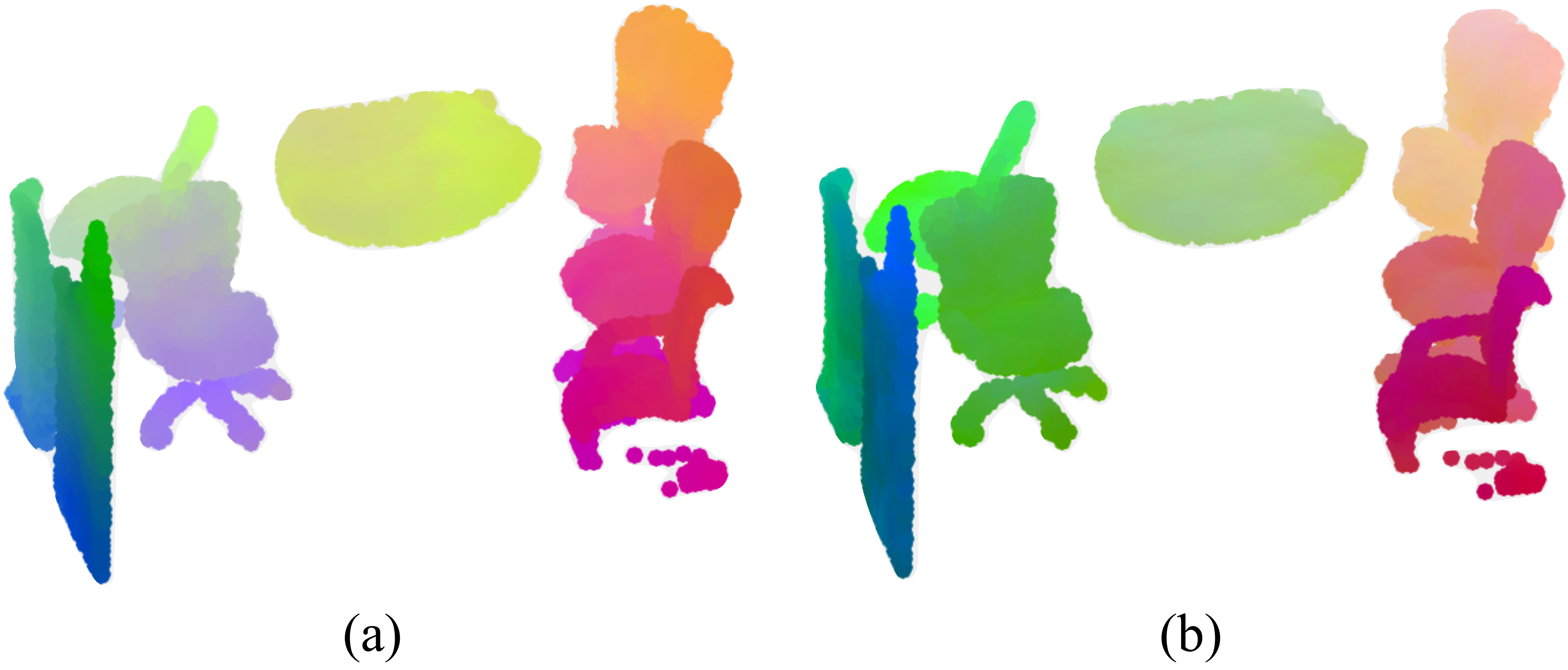}
    \caption{A visualization for the learned instance sensitive features. (a) shows the context features $f_{\hat{c}}$ by first applying PCA to the high dimensional features and then converting the first three dimension into the color map here. (b) shows the predicted object centers. A clearer separation of different instances is shown in (b), which confirms that these features are instance sensitive.}
    \label{fig:context_fea}
    \vspace{-\baselineskip}
\end{figure}

We emphasize that the center prediction network factoring out the location information in the generative model plays a very important role of simplifying the generation task by allowing contexts corresponding to the same object to be encoded with similar context features $f_{\hat{c}}$.
We refer the feature $f_{\hat{c}}$ as \emph{instance sensitive} features and visualize their predictions in Figure~\ref{fig:context_fea}, where a clear difference can be observed among instances. From now on, we overload the symbol $f_{\hat{c}}$ to include the predicted object center $t$. We will also show how these instance sensitive feature could play a role for further proposal refinement and object segmentation in the next section.

\vspace{-\baselineskip}
\paragraph{Losses} GSPN is trained to minimize a multi-task loss function $L_{GSPN}$ defined for each potential object proposal. $L_{GSPN}$ is a summation of five terms including the shape generation loss $L_{gen}$, shape generation per-point confidence loss $L_{e}$, KL loss $L_{KL}$, center prediction loss $L_{center}$, and objectness loss $L_{obj}$. We use chamfer distance between the generated objects $\tilde{x}$ and the ground truth objects $x$ as the generation loss $L_{gen}$, which serves as a surrogate to the negative log likelihood $-\text{log}p_{\theta}(x|z,c)$. To supervise the per-point confidence prediction, we compute the distance from each predicted point to the ground truth object point cloud. Those points with distances smaller than a certain threshold $\epsilon$ are treated as confident predictions and others are unconfident predictions. KL loss essentially enforces the proposal distribution $q_{\phi}(z|x,c)$ and the prior distribution $p_{\theta}(z|c)$ to be similar. Since we have parametrized $q_{\phi}(z|x,c)$ and $p_{\theta}(z|c)$ as $\mathcal{N}(\mu_z^{\prime}, \sigma_z^{\prime 2})$ and $\mathcal{N}(\mu_z, \sigma_z^2)$ respectively through neural networks, the KL loss can be easily computed as:
\vspace{-0.5\baselineskip}
\begin{equation}
L_{KL} = \text{log}\frac{\sigma_z^{\prime}}{\sigma_z}+\frac{\sigma_z^2+(\mu_z-\mu_z^{\prime})^2}{2\sigma_z^{\prime 2}}-0.5
\end{equation}
\noindent
Average binary cross entropy loss is used for $L_e$. Smooth L1 loss \cite{girshick2015fast} is used as the center prediction loss $L_{center}$. $L_{obj}$ is also defined as an average binary cross entropy loss.

\subsection{Region-based PointNet}
\begin{figure*}[t!]
    \centering
    \includegraphics[width=\linewidth]{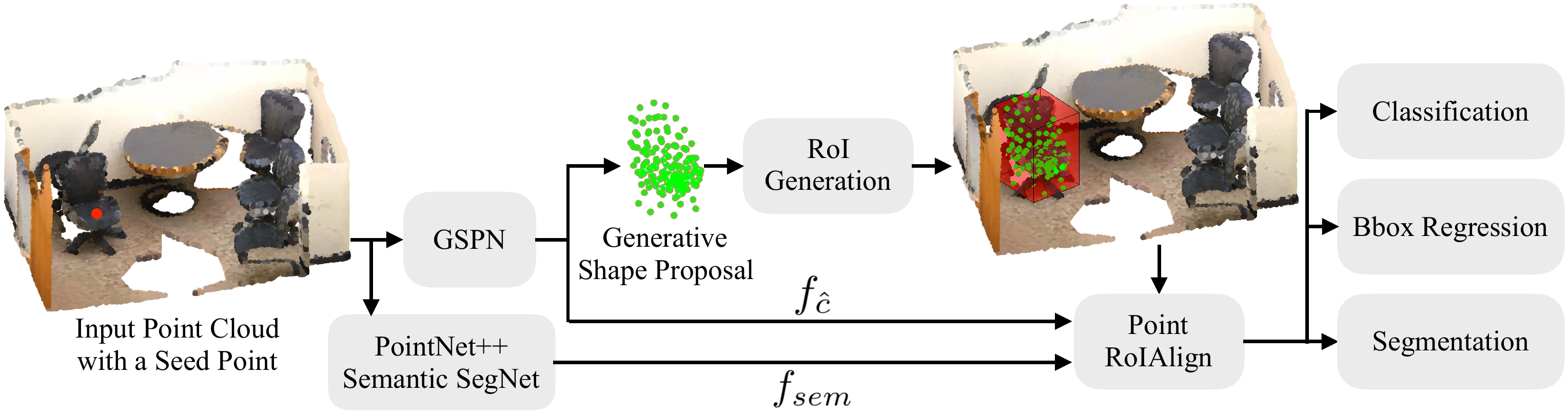}
    \caption{The architecture of R-PointNet. For each seed point in the scene, GSPN would generate a shape proposal along with instance sensitive features $f_{\hat{c}}$. The shape proposal is converted into an axis-aligned 3D bounding box, where Point RoIAlign can be applied to extract RoI features for the final segmentation generation. In addition to instance sensitive feature $f_{\hat{c}}$, semantic features obtained from a pretrained PointNet++ segmentation network are also used in the feature backbone.}
    \label{fig:rpointnet}
    \vspace{-\baselineskip}
\end{figure*}

In the second part of R-PointNet, the object proposals from GSPN are further processed to identify the object class, refine the proposal, and segment the foreground objects in the proposals from the initial point cloud $P$.
We first obtain candidate RoIs by computing axis-aligned bounding boxes from the object proposals (points which confidence score $e$ is greater than 0.5 are only used). Then, from each RoI, our Point RoIAlign layer extracts region features, which are fed through PointNet-based classification, regression, and segmentation sub-networks.
Bounding boxes are refined by predicting the relative center and size changes as done in~\cite{ren2015faster}, and also the segmentation is obtained by predicting a per-point binary mask for each category label similarly with~\cite{he2017mask}.
We visualize the architecture of R-PointNet in Figure~\ref{fig:rpointnet}. R-PointNet is trained to minimize a multi-task loss function defined within each RoI as $L=L_{cls}+L_{box}+L_{mask}$, which is the same as \cite{he2017mask}. Next, we explain design details about R-PointNet.


\vspace{-\baselineskip}
\paragraph{Feature Backbone}
Before computing region features in each RoI, we first augment the context feature $f_{\hat{c}}$ of GSPN with a \emph{semantic} feature $f_{\text{sem}}$ coming from a network pre-trained on a semantic segmentation task. Specifically, we pre-train a PointNet++ classifying each point into object classes with 4 sample-and-group layers and 4 feature interpolation layers. Then, we obtain the semantic feature for each point as a combination of each of the sample-and-group layers outputs in order to capture information at various scales.
Since the point cloud is downsampled after a sample-and-group layer, for covering every point in $P$, we upsample the feature set after each sample-and-group layer through a feature interpolation operation; find the three closest points and interpolate with weights inversely proportional to distances. This allows us to concatenate features with different scales and form hypercolumns \cite{hariharan2015hypercolumns} for points. The concatenation of the context feature $f_{\hat{c}}$ and the semantic feature $f_{\text{sem}}$ builds our feature backbone, and the feature backbones are aggregated in the next Point RoIAlign step. In ablation study~\ref{sec:ablation_study}, we demonstrate that both of the context and semantic features play an important role in obtaining good instance segmentation.

\vspace{-\baselineskip}
\paragraph{Point RoIAlign}
For obtaining a fixed-size feature map within each RoI, our point RoIAlign layer samples $N_{RoI}$ points equipped with a feature vector from $P$. Since the computation of context feature $f_{\hat{c}}$ is very expensive, practically we compute the features just for a set of \emph{seed} points $P_{sub}$, and obtaining features of the sampled points using the feature interpolation step described in the previous paragraph. After point sampling and feature extraction, each RoI is normalized to be a unit cube centered at (0,0,0).

\subsection{Implementation Details}
In all our experiments, we first train GSPN and PointNet++ semantic segmentation network and then fix their weights during R-PointNet training.

\vspace{-\baselineskip}
\paragraph{Training}
While training GSPN, we randomly sample $512$ seed points for each training scene in each mini-batch, which gives us $512$ shape proposals. We use a resolution of $512$ points for these shape proposals. The context around each seed point is represented as a multi-scale cropping of the scene, where the cropping radius is set so that the smallest scale is on a par with the smallest object instance in the scenes and the largest scale roughly covers the largest object. To predict the object center for each seed point, we simply regress the unit direction vector from the seed point to the object center as well as the distance between the two similar to \cite{xiang2017posecnn}. We also adopt KL-annealing \cite{bowman2015generating} to stabilize the GSPN training.

During the training procedure of R-PointNet, we apply non-maximum suppression \cite{girshick2015deformable} on all the object proposals and keep a maximum number of $128$ proposals for training. We select positive and negative RoIs in the same way as \cite{he2017mask} where positive RoIs are those intersecting with ground truth bounding boxes with an IoU greater than $0.5$, and negative RoIs are those with less than $0.5$ IoU with all the ground truth bounding boxes. The ratio between positive and negative RoIs is set to be 1:3.


\vspace{-\baselineskip}
\paragraph{Inference}
During the inference time, we randomly sample $2048$ seed points in each test scene and keep at most $512$ RoIs after non-maximum suppression for RoI classification, refinement, and segmentation. It usually takes $\sim$1s on a Titan XP GPU to consume an entire scene with $\sim$20k points. After obtaining a binary segmentation within each RoI, we project the segmentation mask back to the initial point cloud through the nearest neighbor search. All the points outside the RoI will be excluded from the projection.

\section{Experiment}
Our object proposal module GSPN and the instance segmentation framework R-PointNet are very general and could handle various types of data. To show their effectiveness, we experiment on three different datasets including:
\begin{itemize}
\setlength\itemsep{-0.2em}
    \item ScanNet \cite{dai2017scannet}: This is a large scale scan dataset containing 3D reconstructions of 1613 indoor scenes. Each reconstruction is generated by fusing multiple scans from different views. The scenes are annotated with semantic and instance segmentation masks. They are officially split into 1201 training scenes, 312 validation scenes and 100 test scenes, where ground truth label is only publicly available for the training and validation sets. 
    \item PartNet\cite{mo2018partnet}: This dataset provides fine-grained part instance annotations for 3D objects from ShapeNet \cite{chang2015shapenet}. It covers 24 object categories and the number of part instances per-object ranges from 2 to 220 with an average of 18.
    \item NYUv2 \cite{Silberman:ECCV12}: This dataset contains 1449 RGBD images with 2D semantic instance segmentation annotations. We use the improved annotation from \cite{deng2017amodal}. Partial point cloud could be obtained by lifting the depth channel using the camera information. And we follow the standard train test split.
\end{itemize}
\noindent
We also conduct extensive ablation study to validate different design choices of our framework.

\begin{table*}[t]
\centering
\newcolumntype{Y}{>{\centering\arraybackslash}X}
{\small
\setlength{\tabcolsep}{0.2em}
\renewcommand{\arraystretch}{0.9}
\begin{tabularx}{\textwidth}{>{\Centering}m{1.4cm}|>{\Centering}m{0.8cm}|YYYYYYYYYYYYYYYYYY}
\toprule
  & \small{\textbf{Mean}} & \small{cabi-net} & \small{bed} & \small{chair} & \small{sofa} & \small{table} & \small{door} & \small{win-dow} & \small{book-shelf} & \small{pic-ture} &
  \small{coun-ter} & \small{desk} & \small{cur-tain} & \small{fri-dge} & \scriptsize{shower curtain} & \small{toilet} & \small{sink} & \small{bath-tub} & \small{other} \\
\midrule
    \footnotesize{PMRCNN} & 5.3 & 4.7 & 0.2 & 0.2 & 10.7 & 2.0 & 3.1 & 0.4 & 0.0 & \textbf{18.4} & 0.1 & 0.0 &
    2.0 & 6.5 & 0.0 & 10.9 & 1.4 & 33.3 & 2.1 \\
    \footnotesize{SGPN} & 13.3 & 6.0 & 36.1 & 25.7 & 33.5 & 16.1 & 7.9 & 12.2 & 14.9 & 1.3 & 2.6 & 0.0 & 6.2 & 2.6 & 0.0 & 16.1 & 10.4 & 19.4 & 3.8\\
    Ours & \textbf{30.6} & \textbf{34.8} & \textbf{40.5} & \textbf{58.9} & \textbf{39.6} & \textbf{27.5} & \textbf{28.3} & \textbf{24.5} & \textbf{31.1} & 2.8 & \textbf{5.4} & \textbf{12.6} &
    \textbf{6.8} & \textbf{21.9} & \textbf{21.4} & \textbf{82.1} & \textbf{33.1} & \textbf{50.0} & \textbf{29.0} \\
\bottomrule
\end{tabularx}
}
\vspace{-0.5\baselineskip}
\caption{Instance segmentation results on ScanNet (v2) 3D semantic instance benchmark.}
\vspace{-0.5\baselineskip}
\label{tab:scannet_insseg}
\end{table*}

\begin{figure*}
    \centering
    \includegraphics[width=\linewidth]{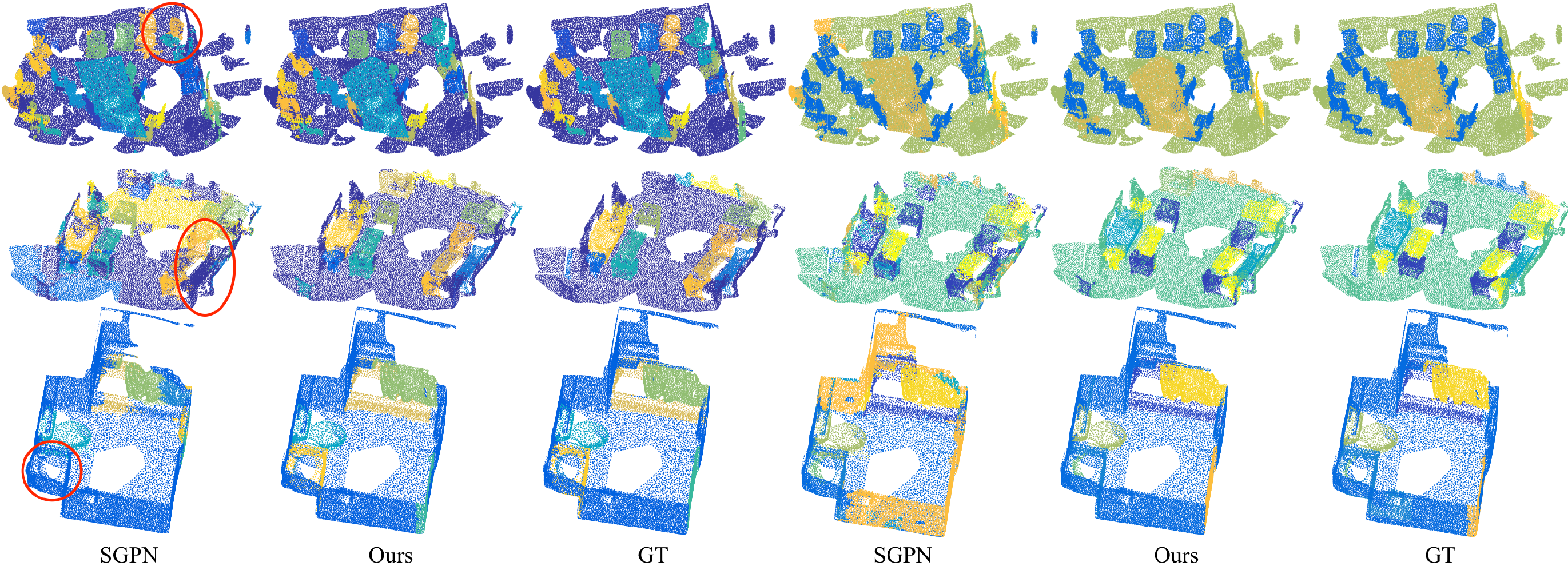}
    \caption{Visualization for ScanNet instance segmentation results. The first three columns show the instance segmentation results where different colors represent different object instances and the last three columns show semantic segmentation results. We highlight SGPN's failure case with red circles in the first column. It is frequent for SGPN to break one object into multiple pieces or miss certain objects.}
    \label{fig:scannet_insseg}
\vspace{-\baselineskip}
\end{figure*}

\subsection{Instance Segmentation on ScanNet}
We first evaluate our approach on ScanNet 3D semantic instance segmentation benchmark, where algorithms are evaluated and compared on 18 common object categories. In this task, colored point clouds are provided as input and the goal is to segment out every instance in the scene belonging to the 18 object categories, along with its semantic label. The list of categories captures a variety of objects from small-scale pictures to large-scale doors, making the task very challenging. The average precision (AP) with an IoU threshold 0.5 is used as the evaluation metric. Different from detection tasks, the IoU here is computed based on segmentations instead of bounding boxes, emphasizing more on detailed understandings. Unlike previous methods \cite{qi2017pointnet,wang2018sgpn}, we do not cut scene into cubes but directly consume the whole scene, which avoids the cube merging process and is much more convenient. We compare our approach with the leading players on the ScanNet (v2) benchmark, including SGPN \cite{wang2018sgpn} and a projected Mask R-CNN (PMRCNN) approach \cite{scannetbenchmark}. SGPN learns to group points from the same instance through a variation of metric learning. PMRCNN first predicts instances in 2D colored images and then project the predictions back to 3D point clouds followed by an aggregation step. To the best of our knowledge, currently these are the only published learning based instance segmentation approaches for 3D point cloud that could handle arbitrary object categories in an indoor environment. We report the results in Table~\ref{tab:scannet_insseg}.

Our R-PointNet outperforms all the previous state-of-the-arts on most of the object categories by a large margin and achieves the leading position on the ScanNet 3D semantic instance segmentation benchmark~\cite{scannetbenchmark}. R-PointNet achieves very high AP for categories with small geometric variations such like toilet since the GSPN only needs to capture a relatively simple object distribution and could generate very high-quality object proposals. For categories requiring strong texture information during segmentation such like window and door, SGPN fails to get good scores since their similarity metric can not effectively encode colors. Our approach achieves much better results on these categories which shows GSPN not only leverages geometry but also the color information while generating object proposals. PMRCNN works in 2D directly instead of in 3D and fails to leverage the 3D information well, leading to very low AP on most of the categories. Interestingly, PMRCNN achieves the best score for the picture category, which is not surprising since pictures lie on 2D surfaces where appearance instead of geometry acts as the key cue for segmentation. 2D based approaches currently are still more capable of learning from the appearance information.

We also show qualitative comparisons between SGPN and our approach in Figure~\ref{fig:scannet_insseg}. SGPN needs to draw a clear boundary in the learned similarity metric space to differentiate object instances, which is not easy. We could observe a lot of object predictions either include a partial object or multiple objects. Compared with SGPN, our GSPN generates proposals with much higher objectness, leading to much better segmentation quality. We also find SGPN having a hard time learning good similarity metric when there are a large number of background points since it purely focuses on learning the semantics and similarity for foreground points and ignores background points during training, which could increase false positive predictions on backgrounds. GSPN, on the other hand, explicitly learns the objectness of each proposal and could easily tell foreground from the background. Therefore it won't be easily influenced by the background points.

\subsection{Part Instance Segmentation on PartNet}
\begin{figure}
    \centering
    \includegraphics[width=\linewidth]{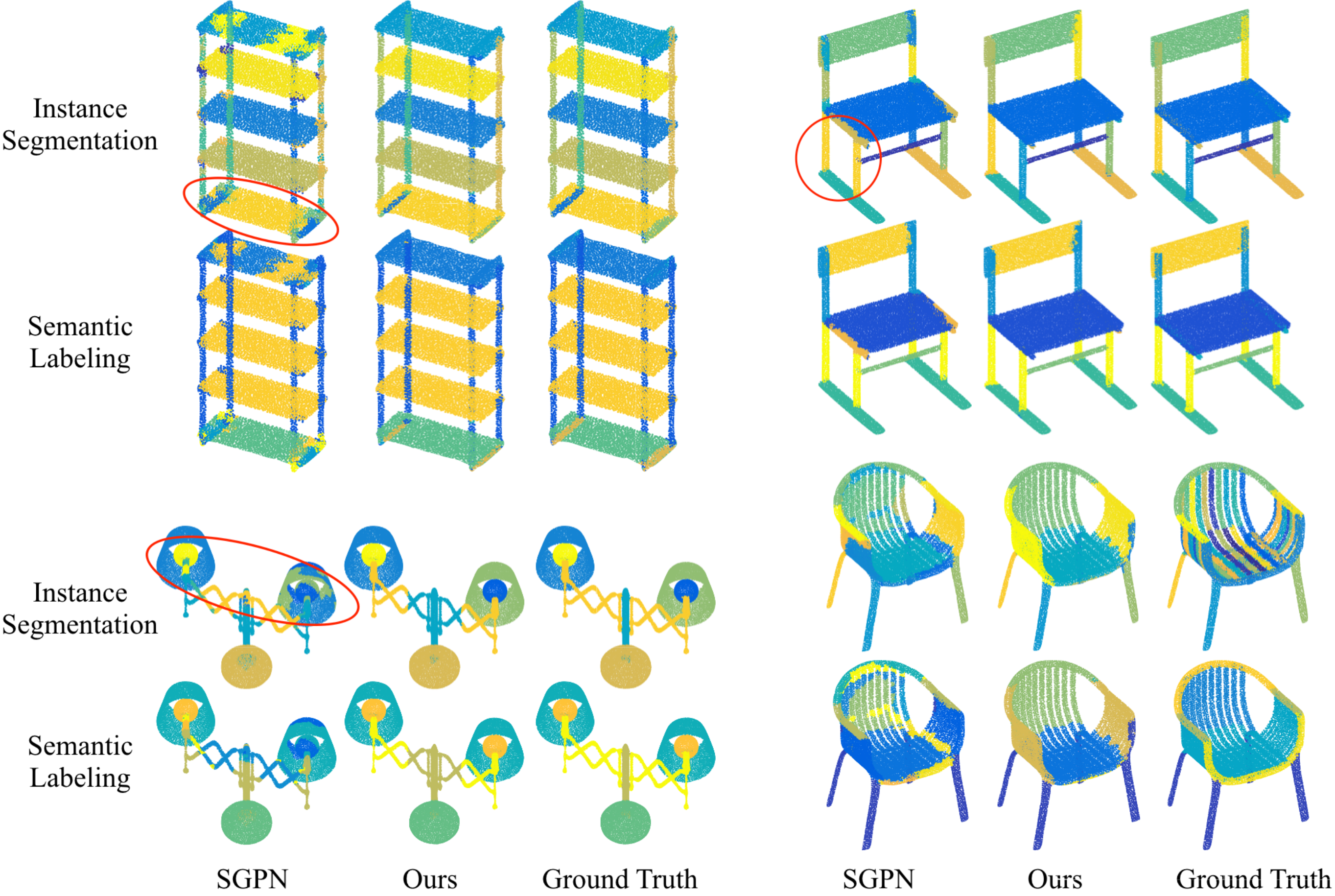}
    \caption{Visualization for part instance segmentation results. As highlighted by the red circles, SGPN does not delineate object parts as good as ours and usually fail to differentiate part instances with the same semantic meaning.}
    \label{fig:partnet_insseg}
\vspace{-1.5\baselineskip}
\end{figure}

Our R-PointNet could not only handle object instance segmentation in indoor scenes, but also part instance segmentation in an object. Different from objects in scenes, object parts are more structured but less separated, e.g. a chair seat always lies above chair legs while closely connecting with them, which introduces new challenges for instance segmentation. \cite{mo2018partnet} introduces PartNet, a large scale fine-grained part annotation dataset, where 3D objects from \cite{chang2015shapenet} are segmented into fine-grained part instances. Different from previous large-scale part annotation dataset \cite{yi2016scalable}, PartNet provides ground truth part instance annotation in addition to semantic labeling and the segmentation granularity is more detailed, making it more suitable for testing instance segmentation approaches. We take the four largest categories from PartNet and evaluate our approach on the semantic part instance segmentation task. Still, we use AP as the evaluation metric with an IoU threshold of 0.5. And we compare with SGPN, which also claims to be able to handle part instance segmentation tasks. We report the qualitative comparison in Table~\ref{tab:partnet_insseg} and also visualize the predictions of both our approach and SGPN in Figure~\ref{fig:partnet_insseg}.

\begin{table}[h]
\vspace{-0.5\baselineskip}
\centering
\newcolumntype{Y}{>{\centering\arraybackslash}X}
{
\setlength{\tabcolsep}{0.2em}
\renewcommand{\arraystretch}{0.9}
\begin{tabularx}{\columnwidth}{Y|Y|Y|Y|Y}
\toprule
    & Chair & Table & Lamp & Storage\\
    \hline
   SGPN & 0.194 & 0.146 & 0.144 & 0.215 \\
   Ours & \textbf{0.268} & \textbf{0.219} & \textbf{0.183} & \textbf{0.267} \\
\bottomrule
\end{tabularx}
}
\caption{Part instance segmentation on PartNet. Our approach outperforms SGPN on all categories.}
\label{tab:partnet_insseg}
\vspace{-0.5\baselineskip}
\end{table}

Our approach outperforms SGPN on all categories by a large margin. As shown in Figure~\ref{fig:partnet_insseg}, our approach could successfully segment part instances with various scales, from small base panels of the storage furniture in the top left corner to large chair seat in the top right corner. Even for part instances enclosed by other parts, e.g. the light bulbs inside the lamp shades in the bottom left corner, we could still segment them out while SGPN fails. SGPN usually groups part instances with the same semantic label. As highlighted by the red circles. We also show a failure case for both our approach and SGPN in the bottom right corner, where each bar on the back, arm, and seat of the chair are treated as an individual part instance. This causes great challenge to both semantic labeling and instance segmentation. Compared with SGPN, we obtain smoother instance segments with much less noise.

\subsection{Instance Segmentation on NYUv2}

\begin{table*}[t]
\centering
\newcolumntype{Y}{>{\centering\arraybackslash}X}
{
\setlength{\tabcolsep}{0.2em}
\renewcommand{\arraystretch}{0.9}
\begin{tabularx}{\textwidth}{>{\Centering}m{1.45cm}|>{\Centering}m{0.8cm}|YYYYYYYYYYYYYYYYYYY}
\toprule
  & \small{\textbf{Mean}} & \small{bath-tub} & \small{bed} & \small{book-shelf} & \small{box} & \small{chair} & \small{coun-ter} & \small{desk} & \small{door} & \small{dres-ser} &
  \small{gar-bin} & \small{lamp} & \small{moni-tor} & \small{night-stand} & \small{pil-low} & \small{sink} & \small{sofa} & \small{table} & \small{TV} & \small{toilet} \\
\midrule
    \footnotesize{SGPN-CNN} & 33.6 & 45.3 & 62.5 & \textbf{43.9} & 0.0 & 45.6 & 40.7 & 30.0 & 20.2 & 42.6 & 8.8 & 28.2 & 15.5 & 43.0 & 30.4 & \textbf{51.4} & \textbf{58.9} & 25.6 & 6.6 & 39.0 \\
    MRCNN & 29.3 & 26.3 & 54.1 & 23.4 & 3.1 & 39.3 & 34.0 & 6.2 & 17.8 & 23.7 & \textbf{23.1} & 31.1 &
    \textbf{35.1} & 25.4 & 26.6 & 36.4 & 47.1 & 21.0 & \textbf{23.3} & 58.8\\
    MRCNN* & 31.5 & 24.7 & \textbf{66.3} & 20.1 & 1.4 & 44.9 & 43.9 & 6.8 & 16.6 & 29.5 & 22.1 & 29.2 &
    29.3 & 36.9 & 34.6 & 37.1 & 48.4 & 26.6 & 21.9 & 58.5\\
    Ours & \textbf{39.3} & \textbf{62.8} & 51.4 & 35.1 & \textbf{11.4} & \textbf{54.6} & \textbf{45.8} & \textbf{38.0} & \textbf{22.9} & \textbf{43.3} & 8.4 & \textbf{36.8} & 18.3 &
    \textbf{58.1} & \textbf{42.0} & 45.4 & 54.8 & \textbf{29.1} & 20.8 & \textbf{67.5}\\
\bottomrule
\end{tabularx}
}
\caption{Instance segmentation results on NYUv2 dataset.}
\label{tab:nyu_insseg}
\vspace{-\baselineskip}
\end{table*}

In this experiment, we focus on colored partial point cloud data lifted from RGBD images. Different from ScanNet, each RGBD image only captures a scene from a single viewpoint, causing a large portion of data missing in the lifted point cloud. This is especially severe in a cluttered indoor environment with heavy occlusions. To show the effectiveness of our approach, we follow the setting of \cite{wang2018sgpn} and conduct instance segmentation for 19 object categories on the colored partial point cloud. AP with an IoU threshold of 0.25 is used as the evaluation metric and again the IoU is computed between predicted segmentations and ground truth segmentations. 

Same as \cite{wang2018sgpn}, to better make use of the color information, we extract features from RGB images directly with a pretrained AlexNet \cite{krizhevsky2012imagenet}. We use features from the conv5 layer and we concatenate the image feature with the PointNet++ semantic segmentation feature to augment $f_{sem}$, which serves as the semantic feature backbone of R-PointNet. The concatenation between the image feature and the point feature happens within the corresponding pixel and 3D point pair, which could be obtained by projecting each 3D point back to 2D and search for its closest pixel. We compare our approach with SGPN-CNN \cite{wang2018sgpn}, the previous state-of-the-art approach on this partial point cloud instance segmentation task. In SGPN-CNN, 2D CNN features are incorporated into SGPN for better leverage of the color information.

We also compare with Mask R-CNN~\cite{he2017mask}, which is initially designed for RGB images. To adapt it for RGBD image processing, we convert the depth channel into an HHA image following \cite{gupta2014learning} and concatenate it with the original RGB image to form a 6-channel input to the network. We initialize the whole network, except for the first convolution layer, with pre-trained coco weights. We carefully train Mask R-CNN following the guideline provided by \cite{maskrcnnwiki}. To be specific, we first freeze the feature backbone, train the conv1, and heads for 20 epochs with a learning rate $0.001$. And then Resnet 4+ layers are finetuned with another 15 epochs using lower learning rate ($0.0001$). Finally Resnet 3+ layers are open to train for 15 epochs with a learning rate $0.00001$. Due to the small size of training data set (only 795 images), data augmentations (Fliplr \&  Random Rotation \& Gamma Adjustment), high weight decay (0.01) and simple architecture (Resnet-50) are applied to avoid severely overfitting.
This approach is called MRCNN*. We also train Mask R-CNN on the RGB images only to analyze the effectiveness of 3D learning.

While conducting the comparison with \cite{wang2018sgpn}, we found issues within the authors' evaluation protocol. With the help of the authors, we re-evaluate SGPN-CNN and report the quantitative comparisons in Table~\ref{tab:nyu_insseg}.

Our R-PointNet outperforms SGPN-CNN, Mask R-CNN and Mask R-CNN* with respect to mAP and provides the best results on 12/19 classes. Even on partial point cloud with severe data missing, R-PointNet still captures the shape prior, generates object proposals and predicts final segmentations reasonably well. R-PointNet achieves higher scores on categories with small geometric variations such like bathtub and toilet, whose shape distributions are relatively easier for GSPN to capture. 


For shape categories with strong appearance signatures but weak geometric features, such as monitor, Mask R-CNN achieves the best performance. This indicates our way of using color information is not as effective as Mask R-CNN. Introducing depth information to Mask R-CNN does not improve its performance dramatically, even on categories with a strong geometric feature which could be easily segmented out by R-PointNet such like bathtub. This justifies the necessity of 3D learning while dealing with RGBD images.



\subsection{Ablation Study}
\label{sec:ablation_study}
To validate various design choices of ours, we conduct ablation study on the ScanNet (v2) validation set and discuss the details below.

\begin{figure}
    \centering
    \includegraphics[width=\linewidth]{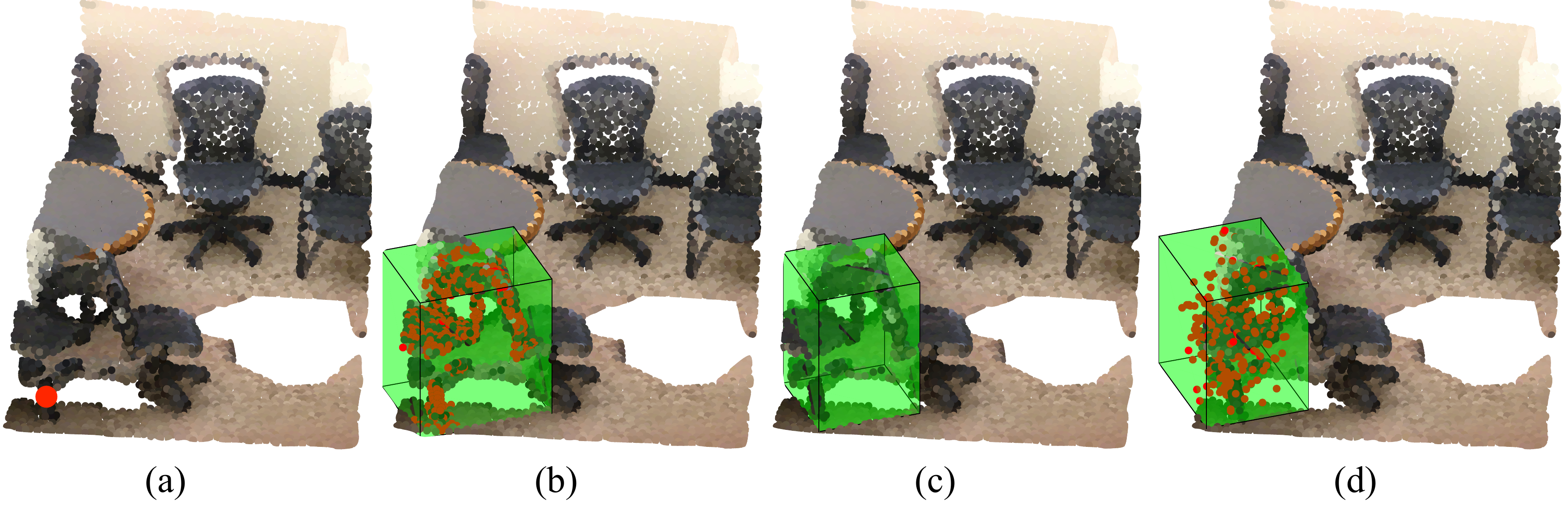}
    \caption{Visualization of object proposals with the induced 3D bounding boxes generated by different approaches. (a) shows the input point cloud and the red point on the chair leg serves as the seed point. We show proposals generated by (b) binary segmentation based object proposal, (c) direct bounding box regression, (d) GSPN. The 3D bounding boxes given by (b) and (c) include other objects while GSPN generates a faithful approximation to the underlying object thus successfully avoids including other objects.}
    \label{fig:abl_proposal}
    \vspace{-1.5\baselineskip}
\end{figure}

\vspace{-\baselineskip}
\paragraph{Comparison with Other 3D Proposal Approaches}
We conduct ablation study by replacing GSPN with other types of object proposal networks. To be specific, we implemented two alternatives. One is to directly regress 3D bounding box for each object, mimicking the Mask R-CNN. For each seed point in the 3D space, we define 27 axis-aligned anchor boxes associated with three scales and 9 aspect ratios. We design a proposal network, which is essentially an MLP, for bounding box regression. It takes in the context features $f_{\hat{c}}$ and directly regress center shift, box deltas and objectness score for each anchor box, which is used for anchor box refinement. Then we apply non-maximum suppression and point RoI align to the refined anchor boxes and feed them into the final classification, box regression, and segmentation heads. The other way is to conduct a binary segmentation within a specific context centered at each seed point and convert the resulting segmentation into axis-aligned bounding box for object proposals. We choose the largest context $c_3$ in our case to guarantee that the largest object in the scene could be fully included by the context. We compare the RoI generated by GSPN with the above two alternatives through two evaluation metrics. One is the mean 3D IoU (mIoU) between the proposal bounding boxes and ground truth bounding boxes at all seed points. The other is the final instance segmentation mAP. For direct bounding box regression approach, we select the refined anchor box with the highest objectness score for mIoU computation. We report the quantitative comparison in Table~\ref{tab:abl_proposal} and visualize proposals from different approaches in Figure~\ref{fig:abl_proposal}.

Our GSPN achieves the best performance with respect to both evaluation metrics. It generates better object proposals more similar to the underlying objects, which indeed improves the final segmentation performance. Both binary segmentation based object proposal and bounding box regression based object proposals could generate boxes covering partial or multiple objects, influencing their usage for the downstream segmentation task.

\begin{table}[h]
\centering
\newcolumntype{Y}{>{\centering\arraybackslash}X}
{\small
\setlength{\tabcolsep}{0.2em}
\renewcommand{\arraystretch}{0.9}
\begin{tabularx}{\columnwidth}{Y|Y|Y|Y|Y}
\toprule
     & mIoU & AP & AP@0.5 & AP@0.25\\
\midrule
   Bbox Reg & 0.514 & 15.8 & 33.1 & 51.3 \\
   Binary Seg & 0.543 & 14.9 &  30.9 & 47.7 \\
   Ours & \textbf{0.581} & \textbf{19.3} & \textbf{37.8} & \textbf{53.4} \\
\bottomrule
\end{tabularx}
}
\caption{Evaluation of different 3D proposal approaches. Compared with straightforward bounding box regression and binary segmentation based bounding box proposal, our GSPN not only generates object proposals overlapping more with the ground truth objects, but also largely improves the final segmentation mAP.}
\label{tab:abl_proposal}
\vspace{-\baselineskip}
\end{table}

\paragraph{Generative Model Design}
In GSPN, we use a variation of CVAE which takes multi-scale context around each seed point as input and predicts the corresponding object center before the generation process. To validate our design choices, we experiment with three additional settings: replacing CVAE with naive encoder-decoder structure (E-D), using a single scale context as input, and removing the center prediction network from GSPN. In the naive encoder-decoder structure, we encode the multi-scale context into a latent code and directly decode the object instance from the code. For the second setting, we choose the largest context as input to guarantee the largest objects in the scene can be roughly included. We use chamfer distance (CD) and mIoU as our evaluation metrics, where CD is computed between the generated shape and the ground truth shape, and mIoU is computed between the induced axis-aligned bounding boxes from the shapes. The comparison is reported in Table~\ref{tab:abl_proposal}. 

When replacing CVAE with E-D, we observe difficulties for the network to produce good shape reconstructions. We conjecture the instance guidance in CVAE training makes such a difference. The recognition network encodes the ground truth instances into a proposal distribution, which provides strong guidance to the prior network to learn a semantic meaningful and compact prior shape distribution. We also find using single context fails to encode small objects well, leading to worse object proposals. The experiment also shows it is important to explicitly predict object center to learn the generation process in a normalized space, otherwise the reconstruction quality will be influenced.

\begin{table}[h]
\centering
\newcolumntype{Y}{>{\centering\arraybackslash}X}
{\small
\setlength{\tabcolsep}{0.2em}
\renewcommand{\arraystretch}{0.9}
\begin{tabularx}{\columnwidth}{Y|Y|Y|Y|>{\Centering}m{2cm}}
\toprule
    & Ours & E-D & 1-Context & No Center Pred\\
\midrule
   CD & \textbf{0.0450} & 0.0532 & 0.0524 & 0.0571 \\
   mIoU & \textbf{0.581} & 0.408 & 0.486 & 0.409 \\
\bottomrule
\end{tabularx}
}
\caption{Evaluation of different generative model designs. Using CVAE with multi-scale context inputs, along with a center prediction network for translation factorization, gives us the best proposal generation quality.}
\label{tab:my_label}
\end{table}

\vspace{-0.2cm}
\paragraph{Choices of Feature Backbone}
We use a combination of instance sensitive context feature $f_{\hat{c}}$ and semantic feature $f_{sem}$ as the feature backbone, which both play an important role in achieving good instance segmentation. We validate their importance by removing each of them and evaluating the influence to the final segmentation mAP. The results are shown in Table~\ref{tab:feat_abl}. It can be seen that removing either of them from the backbone will cause a performance degeneration. This confirms that the instance sensitive feature GSPN learns is complementary to the semantic feature $f_{sem}$, and using both is important. In addition, we also remove the pretraining step for $f_{sem}$ and train the pointnet++ semantic segmentation network in an end-to-end fashion with R-PointNet. We observe a performance drop as well as is shown in Table~\ref{tab:feat_abl}.

\begin{table}[h]
\centering
\newcolumntype{Y}{>{\centering\arraybackslash}X}
{\small
\setlength{\tabcolsep}{0.2em}
\renewcommand{\arraystretch}{0.9}
\begin{tabularx}{\columnwidth}{>{\Centering}m{2.4cm}|Y|Y|Y}
\toprule
    & AP & AP@0.5 & AP@0.25\\
\midrule
    w/o $f_{\hat{c}}$ & 0.178 & 0.349 & 0.515 \\
    w/o $f_{sem}$ & 0.161 & 0.319 & 0.477 \\
    w/o pretraining & 0.180 & 0.364 & 0.517 \\
    Ours & \textbf{0.191} & \textbf{0.376} & \textbf{0.541}\\
\bottomrule
\end{tabularx}
  }
\caption{Comparison of different choices for the feature backbone. Both context feature $f_{\hat{c}}$ and semantic feature $f_{sem}$ play important roles in our feature backbone. We also find pretraining the semantic feature with a semantic segmentation task improves the segmentation performance.}
\label{tab:feat_abl}
\vspace{-\baselineskip}
\end{table}


\subsection{Additional Visualizations}
\label{sec:vis}
\begin{figure*}
    \centering
    \includegraphics[width=\linewidth]{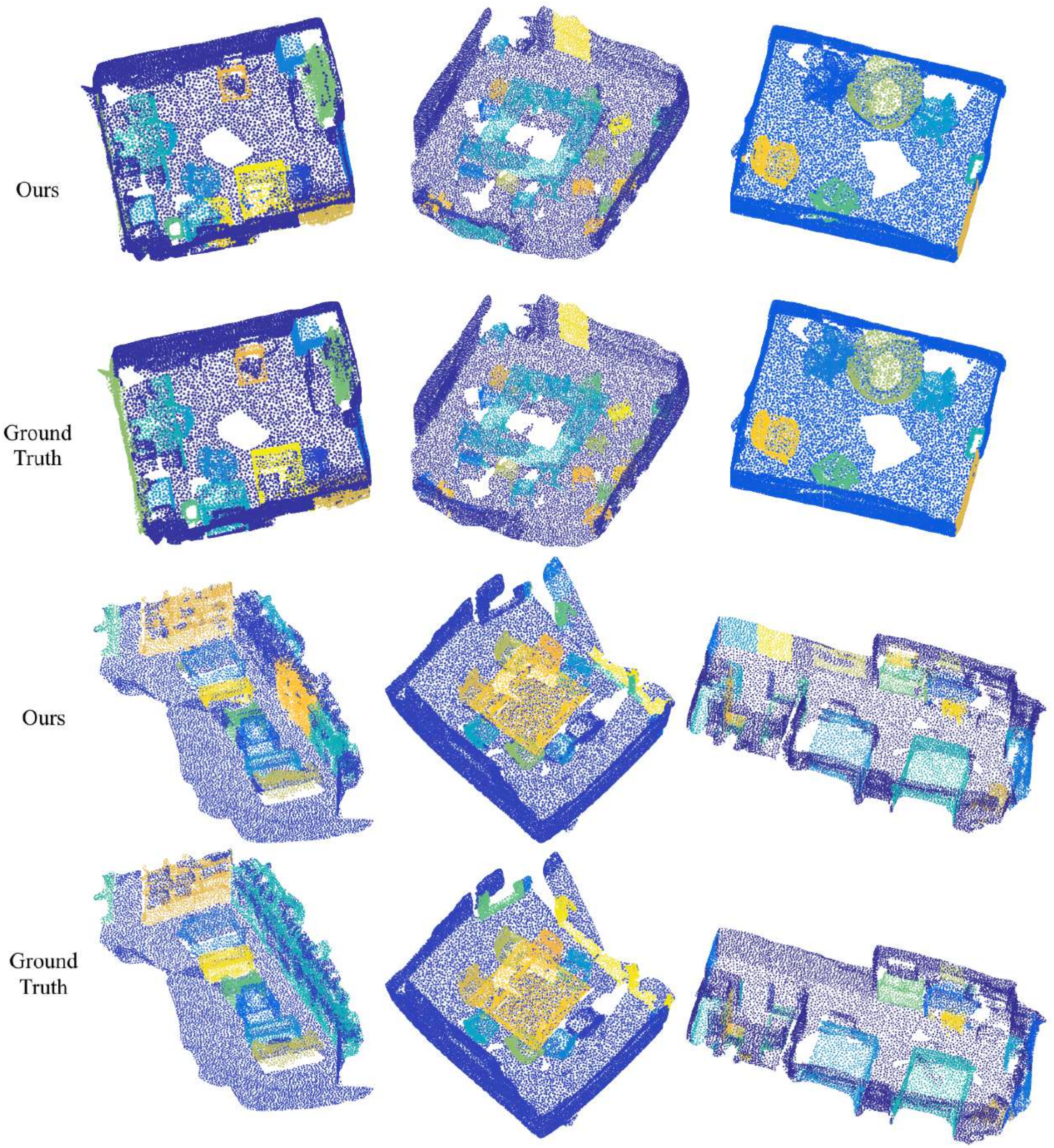}
    \caption{Visualization for ScanNet instance segmentation results. Different colors indicate different instances.}
    \label{fig:scannetseg_supp}
\vspace{-\baselineskip}
\end{figure*}

\begin{figure*}
    \centering
    \includegraphics[width=\linewidth]{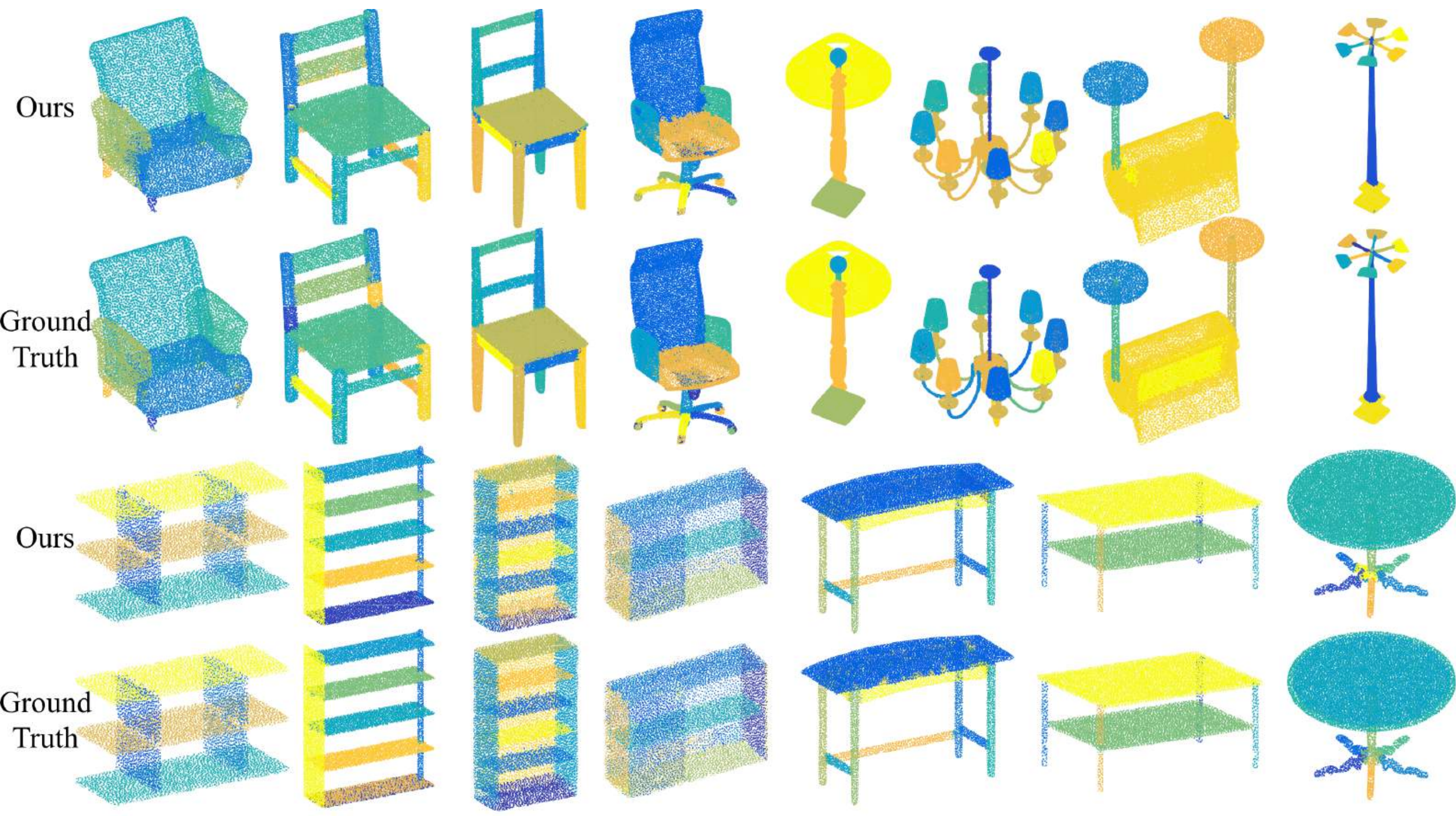}
    \caption{Visualization for PartNet instance segmentation results. Different colors indicate different instances.}
    \label{fig:partnetseg_supp}
\vspace{-\baselineskip}
\end{figure*}

\begin{figure*}
    \centering
    \includegraphics[width=\linewidth]{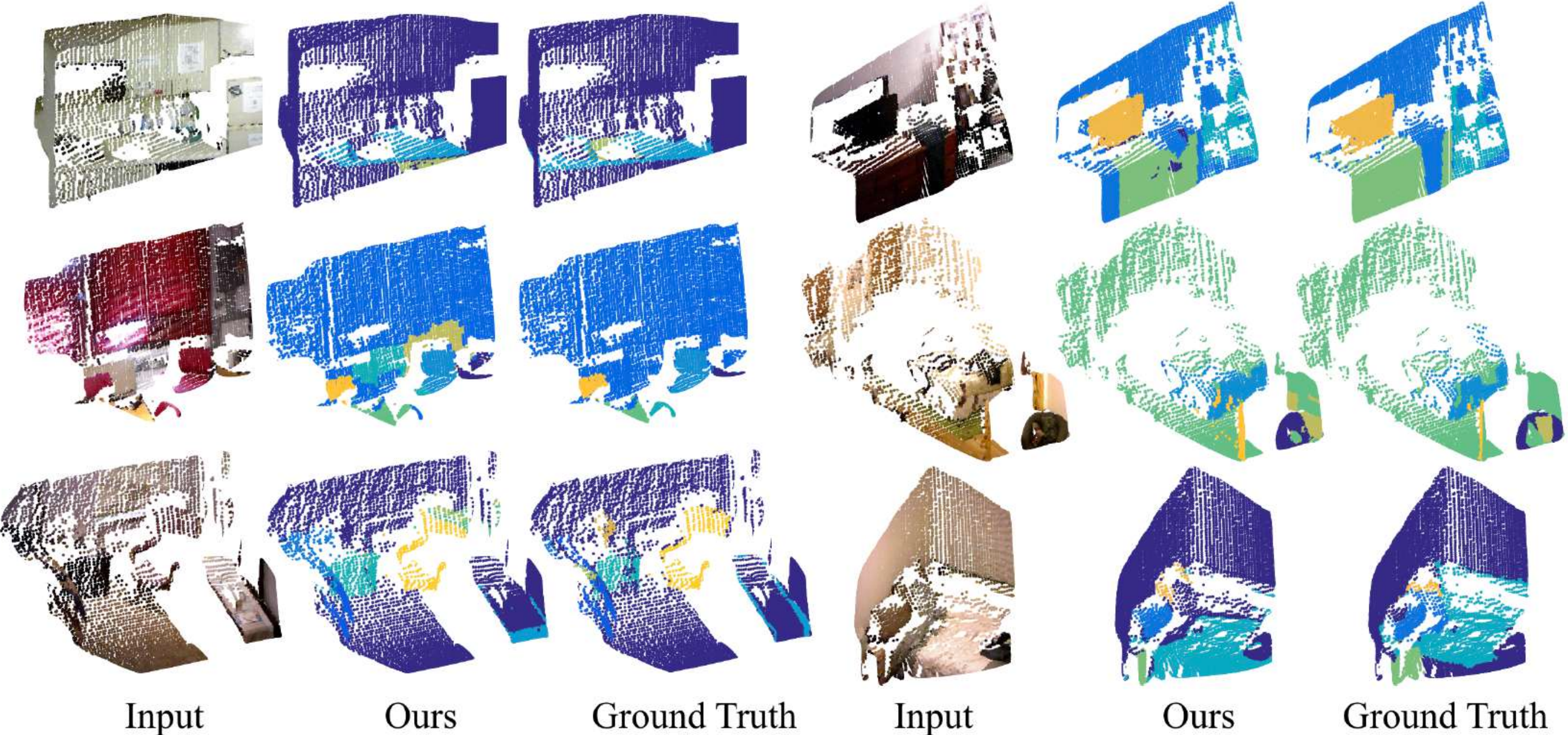}
    \caption{Visualization for NYUv2 instance segmentation results. Different colors indicate different instances.}
    \label{fig:nyuseg_supp}
\vspace{-\baselineskip}
\end{figure*}

To better understand the result quality that can be achieved by our approach, we provide more visualizations in this section. Specifically, we show instance segmentation results on ScanNet, PartNet and NYUv2 in Figure~\ref{fig:scannetseg_supp}, Figure~\ref{fig:partnetseg_supp} and Figure~\ref{fig:nyuseg_supp} respectively.

\section{Conclusion}
We present GSPN, a novel object proposal network for instance segmentation in 3D point cloud data. GSPN generates good quality object proposals with high objectness, which could greatly boost the performance of an instance segmentation framework. We demonstrate how GSPN could be incorporated into a novel 3D instance segmentation framework: R-PointNet, and achieve state-of-the-art performance on several instance segmentation benchmarks.


\paragraph{Acknowledgement} This work was supported by NSF grants CHS-1528025 and IIS-1763268, a Vannevar Bush faculty fellowship, a Google Focused Research Award, and a gift from Amazon AWS.

{\small
\bibliographystyle{ieee}
\bibliography{egbib}
}

\appendix
\appendix 

\section{Architecture Details}
\label{sec:arc_details}
In this section, we provide architecture details about GSPN and Region-based PointNet (R-PointNet)

\begin{table}[h]
    \centering
    \begin{tabular}{c|c}
    \toprule
        Sub-Networks & Architecture \\
    \midrule
        Center \\Prediction Net & SA(2048, 0.2, [32, 32, 64])  \\
        & SA(512, 0.4, [64, 64, 128]) \\
        & SA(128, 0.8, [128, 128, 256]) \\
        & SA(32, 1.6, [256, 256, 512]) \\
        & FP(256, 256) \\
        & FP(256, 256) \\
        & FP(256, 128) \\
        & FP(128, 128, 128) \\
        \midrule
        Prior Net & MLP([64, 128, 256]) (For context) \\
        & MLP([256, 512, 512])  (After concat)\\
        \midrule
        Recognition Net & MLP([64, 256, 512, 256])  \\
        & MLP([256, 512, 512]) \\
        \midrule
        Generation Net & Deconv(512, [3,3], [1,1]) \\
        & Deconv(256, [3,3], [2,2]) \\
        & Deconv(128, [4,4], [2,2]) \\
        & Deconv(3, [1,1], [1,1]) \\
        & FC([512, 512, 256*3]) \\
    \bottomrule 
    \end{tabular}
    \caption{Architecture details of GSPN.}
    \label{tab:gspn_net}
\end{table}

In GSPN, we have a center prediction network, a prior network, a recognition network and a generation network. We use PointNet/PointNet++ as their backbones. Following the same notations in PointNet++, SA$(K, r, [l_1, ..., l_d])$ is a set abstraction (SA) layer with $K$ local regions of ball radius $r$. The SA layer uses PointNet of $d$ 1 x 1 $conv$ layers with output channels $l_1, ..., l_d$ respectively. FP$(l_1, ..., l_d)$ is a feature propagation (FP) layer with $d$ 1 x 1 $conv$ layers, whose output channels are $l_1, ..., l_d$. Deconv$(C, [h, w], [s_1, s_2])$ means a deconv layer with $C$ output channels, a kernel size of $[h, w]$ and a stride of $[s_1, s_2]$. MLP$([l_1, ..., l_d])$ indicates several multi-layer perceptrons (MLP) with output channels $l_1, ..., l_d$. FC$([l_1, ..., l_d])$ is the same as MLP. We report the details of the network design in Table \ref{tab:gspn_net}. The center prediction network is essentially a PointNet++. The prior network takes three contexts as input and uses three PointNets to encode each context. The parameters of the MLP used within each PointNet are shown in the first list. Then their features are concatenated and several MLPs are applied to transform the aggregated features to get the mean \& variance of the latent variable $z$. In the recognition network, the first list of MLPs is used to extract shape features and the second list of MLPs is used to output the mean \& variance of the latent variable $z$. In the generation network, we use both $deconv$ and $fc$ layers to generate shape proposals. This two branches (deconv and fc) generate parts of the whole point set independently which are combined together in the end.

\begin{table}[h]
    \centering
    \begin{tabular}{c|c|c}
    \toprule
        Configurations & Train & Inference \\
    \midrule
        num\_sample & 512 & 2048 \\
        spn\_pre\_nms\_limit & 192 & 1536 \\
        spn\_nms\_max\_size & 128 & 384 \\
        spn\_iou\_threshold & 0.5 & 0.5 \\
        num\_point\_ins\_mask & 256 & 1024 \\
        train\_rois\_per\_image & 64 & - \\
        detection\_min\_confidence & 0.5 & 0.5 \\
        detection\_max\_instances & 100 & 100 \\
    \bottomrule 
    \end{tabular}
    \caption{Main configuration parameters used during train and inference.}
    \label{tab:config}
\end{table}

The R-PointNet consists of three heads: a classification head, a segmentation head and a bounding box refinement head. For the classification head and the bounding box refinement head, we first use an MLP with feature dimensions (128, 256, 512) to transform the input features. Then after max-pooling, we apply several fully-connected layers with output dimensions (256, 256, num\_category) and (256, 256, num\_category*6) to get the classification scores and the bounding box updates, respectively. For the segmentation head, we choose to use a small PointNet segmentation architecture with MLP([64, 64]) for local feature extraction, MLP([64, 128, 512]) \& Max-pooling for global feature extraction and 1x1 conv(256, 256, num\_categroy) for segmentation label prediction. we predict one segment for each category and the weights are updated only based on the prediction for the ground truth category during training. During the inference time, the predicted RoIs are refined based on the bounding box refinement head, which then goes through Point RoIAlign to generate RoI features for the segmentation head.

In addition, we provide various configuration parameters in Table~\ref{tab:config}. ``num\_sample'' represents the number of seed points we used for shape proposal. ``spn\_pre\_nms\_limit'' represents the object proposals we keep after GSPN by filtering out proposals with low objectness scores. ``spn\_nms\_max\_size'' is the maximum number of object proposals we keep after a non-maximum suppression operation following GSPN. ``spn\_iou\_threshold'' is the 3D IoU threshold we used for the non-maximum suppression operation. ``num\_point\_ins\_mask'' is the number of points in each of our generated shape proposals. ``train\_rois\_per\_image'' is the maximum number of RoIs we use for training within each image in each mini-batch. ``detection\_min\_confidence'' is the confidence threshold we use during the inference time, where detections with confidence scores lower than this threshold are filtered out. ``detection\_max\_instances'' is the maximum number of instances we detection from a single scene or object.

\end{document}